\documentclass[journal]{IEEEtran}
\usepackage{orcidlink}
\usepackage{gensymb}
\usepackage{amsmath}
\usepackage{multirow}
\usepackage[skip=5pt]{caption}
\usepackage{enumitem}

\usepackage{subcaption}
\PassOptionsToPackage{hidelinks}{hyperref}
\usepackage{url}
\usepackage[hidelinks]{hyperref}
\hypersetup{breaklinks=true}
\usepackage{makecell}
\usepackage{wrapfig}
\usepackage{xcolor, soul}
\sethlcolor{red}

\ifCLASSINFOpdf
\else
\fi

\begin{document}
%
\title{Efficient Transformer-based Hyper-parameter Optimization for Resource-constrained IoT Environments}

%
%
%

\author{Ibrahim~Shaer \orcidlink{0000-0002-2723-4662}\,~\IEEEmembership{Member,~IEEE,}
    Soodeh~Nikan,~\IEEEmembership{Senior Member,~IEEE,}
        and~Abdallah~Shami \orcidlink{0000-0003-2887-0350},~\IEEEmembership{Senior Member,~IEEE}
\thanks{Ibrahim Shaer, Soodeh Nikan and Abdallah Shami are with the Department of Electrical and Computer Engineering, University of Western Ontario, London, ON N6A 3K7, Canada (emails: ishaer@uwo.ca, snikan@uwo.ca, abdallah.shami@uwo.ca)}      
        }

%
%

\markboth{}%
{Shell \MakeLowercase{\textit{et al.}}: Bare Demo of IEEEtran.cls for IEEE Journals}
%



\maketitle

\begin{abstract}
The hyper-parameter optimization (HPO) process is imperative for finding the best-performing Convolutional Neural Networks (CNNs). The automation process of HPO is characterized by its sizable computational footprint and its lack of transparency; both important factors in a resource-constrained Internet of Things (IoT) environment. In this paper, we address these problems by proposing a novel approach that combines transformer architecture and actor-critic Reinforcement Learning (RL) model, \textit{TRL-HPO}, equipped with multi-headed attention that enables parallelization and progressive generation of layers. These assumptions are founded empirically by evaluating \textit{TRL-HPO} on the MNIST dataset and comparing it with state-of-the-art approaches that build CNN models from scratch. The results show that \textit{TRL-HPO} outperforms the classification results of these approaches by 6.8\% within the same time frame, demonstrating the efficiency of \textit{TRL-HPO} for the HPO process. The analysis of the results identifies the main culprit for performance degradation attributed to stacking fully connected layers. This paper identifies new avenues for improving RL-based HPO processes in resource-constrained environments.

\end{abstract}


%
\section{Introduction}

Convolutional Neural Networks (CNNs) are the staple implementation of neural networks (NNs) for image classification and object detection. The progress in this field is attributed to the increased complexity of CNN architectures in terms of the type of connections among different layers and their depth, which increased the computational demands \cite{shaer2023corrfl}. Therefore, the accuracy of these models depends on the number and the type of layers, the connections between these layers, and the parameters assigned to each layer \cite{elsken2019neural}. Other important factors dictating the design of CNNs include training time, inference time, and energy consumption, emphasized in the resource-constrained Internet of Things (IoT) environment. The number of possible choices for designing CNNs is extremely large, which promoted the automated CNN architecture search, achieved through Neural Architecture Search (NAS).

The NAS field has recently seen great progress, due to the incorporation of Reinforcement Learning (RL) agents to search for the best CNN configurations. The RL's appeal stems from its generalizability of hyper-parameter (HP) combinations via function approximation and the trial-and-error approach, which can reduce the computational demand of NAS \cite{baymurzina2022review}. This field is dominated by the literature that either builds CNNs from scratch using basic NN layers or a pre-defined collection of these layers or optimizes the HPs of an already-existing CNN. 

Autonomous vehicles (AV) are part of the envisioned IoT applications that utilize edge servers of limited computational resources. The realization of AVs heavily depends on image classification models deployed on these servers. Their disparate computational capabilities are prohibitive for the prolonged execution of large models. Model partitioning is key to resolving these limitations toward fulfilling the promise of AVs \cite{xu2023cnn}. This requires gathering insights into the contribution of each CNN model's layers to the classification results. In current implementations, this vision is stalled by multiple oversights. The sequential nature of model generation results from the dependence between layer combinations, which can be reflected by the recurrent structures of Recurrent Neural Networks (RNNs). Therefore, RNN-based controllers constructing CNN models dominated the RL-based NAS field \cite{zoph2016neural, baker2016designing}. The one-at-a-time processing of inputs inhibits their parallelization, resulting in their insurmountable computational footprint that is ill-fitted for the resource-scarce vehicular environment. On a different note, these controllers are built to receive a reward when a terminal condition is encountered, masking the contribution of each layer to prediction results. Within the constraints of computational resources and model partitioning requirements, the state-of-the-art (SOTA) approaches are limited by their impracticability and lack of transparency, hindering their deployment in real-world environments. 

\begin{figure*}
    \centering
    \includegraphics[scale=0.45]{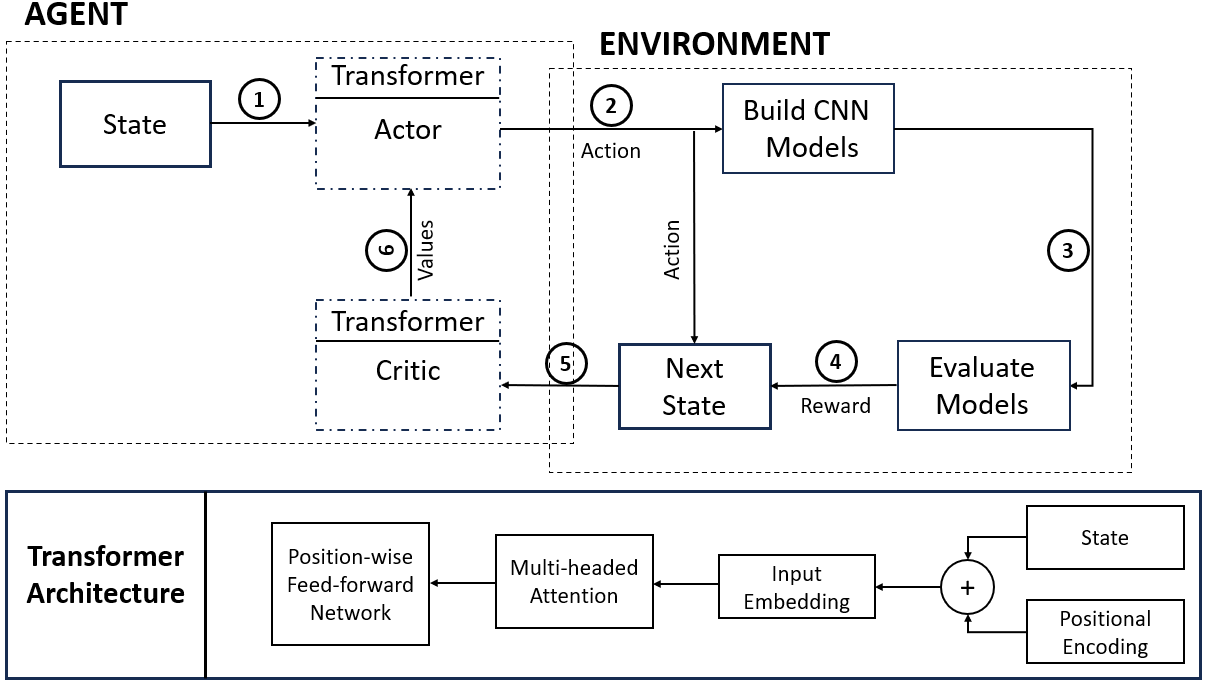}
    \caption{\textit{TRL-HPO} framework}
    \label{fig:trhlpo_framework}
\end{figure*}

This paper presents a Transformer-based Reinforcement Learning Hyper-parameter Optimization (\textit{TRL-HPO}) model that alleviates the shortcomings of the current methods. The \textit{TRL-HPO} addresses the transparency and long-computational times of RL-based solutions with competitive performances. The Transformer architecture overshadows the RNN-based methods by integrating the multi-headed self-attention (MHSA) mechanism that enables parallelization \cite{vaswani2017attention}, addressing the long computational times. The \textit{TRL-HPO} controller equipped with the attention mechanism allows us to gain insights from the layer-generation process, enhancing its interpretability. The reward is produced with every generated layer to showcase improvement instead of waiting for termination conditions to unfold, a condition matching CNN partitioning requirements, as depicted in Figure \ref{fig:trhlpo_framework}. The contributions of this paper are as follows: 

\begin{itemize}
    \item Propose a novel Hyper-parameter Optimization (HPO) process named \textit{TRL-HPO} that is the first to combine transformer architecture and actor-critic (AC) RL; 
    \item Enhance the CNN's model performance generated within a shorter period compared to SOTA approaches using the \textit{TRL-HPO} process;
    \item Improve the transparency of the CNN model generation process by examining the attention-reward affinities and their layer combinations;
    \item Create open challenges related to RL-based HPO processes, including \textit{TRL-HPO}, for researchers to address.
\end{itemize}

The rest of the paper is organized as follows. Section II presents the related work and its limitations. Section III details the proposed framework. Section IV explains the implementation details and evaluation criteria. Section V analyzes the obtained results. Section VI concludes the paper. 
 
\section{Related Work}
The field of hyper-parameter optimization (HPO) is an important research topic in Machine Learning (ML) with practical implementations in real-world environments that enhance the performance of Deep Neural Networks (DNNs). 

Bayesian Optimization (BO), Evolutionary Search (ES) algorithms, and RL agents are the three main tools for HPO implementation. BO and ES methods are limited by their assumptions \cite{elsken2019neural} and lack of generality \cite{yang2020hyperparameter}, which favors RL techniques. The works of Baker \textit{et al.} \cite{baker2016designing} and Zoph \textit{et al.} \cite{zoph2016neural} are the first to propose incorporating RL methods into generating CNN models. The former work utilized RNN-based controllers using an off-policy RL algorithm to sample CNN architectures while the latter utilized a value-based Q-learning approach. The results of these seminal works highlighted the trade-off between the running time of HPO methods and the accuracy of the obtained models. 

The computational footprint of RL implementations for HPO promoted the utilization of multi-agent RL that shrinks the state space of each agent. The work of Neary \cite{neary2018automatic} uses multiple agents to optimize HPs of the CNNs built from scratch, whereby a master agent orchestrates the synchronization between the other agents outputting HPs. On the other hand, the work in \cite{iranfar2021multiagent} assigns each agent to optimize the HPs of an already-existing CNN layer, such that the dependence in HP space is mapped using a shared Q-table between consecutive layers. While all of these studies focus on a single objective, the works presented by Hsu \textit{et al.} \cite{hsu2018monas} and Tan \textit{et al.} \cite{tan2019mnasnet} incorporate multiple objectives in the reward function formulation. MONAS \cite{hsu2018monas} considers the energy and accuracy constraints, whereas MNASNet \cite{tan2019mnasnet} integrates the inference latency of developed models. 

The shortcomings of RL integration into the HPO process hinder their deployment in IoT environments. First, the adopted models lack the transparency that shows the interdependence between different layers. Second, the long convergence times are prohibitive for deploying these models in resource-constrained environments. Third, some works focus on optimizing the HPs of a specific layer, reducing the state space at the expense of the layer's diversity. Lastly, many works include prior knowledge in the layer-generation process, such as the addition of dropout layers \cite{zoph2016neural, baker2016designing}. This paper addresses these limitations by proposing a transformer-based RL controller and the reward function formulation. The MHSA facilitates the training process and adds transparency to the model generation by analyzing the attention values. On the other hand, the reward function reflects the contribution of each layer to enhancing the classification results, which favors CNN model partitioning. 

\section{Methodology: \textit{TRL-HPO}}
This section explains the building blocks of \textit{TRL-HPO}, including the transformer and RL's AC architectures, which are shown in Figure \ref{fig:trhlpo_framework}, and the motivation for this combination toward a more efficient and transparent HPO process.                                                       

\subsection{Transformer}
The vanilla transformer is a sequence-to-sequence model, which avoids the recurrence structures of RNNs by integrating the innovative self-attention mechanism \cite{vaswani2017attention}. This architecture enables parallelization accelerating the training of transformer models. The building blocks of transformers include attention mechanism, multi-headed attention (MHSA), position-wise feed-forward network, and positional encoding (PE). The transformers follow the encoder-decoder structure thereby each of these structures is formed by stacking combinations of these identical layers, as depicted in Figure \ref{fig:trhlpo_framework}. 

The PE computes the sequence's order and is added to the input embedding of the encoder and decoder stacks so that the order is incorporated into the input and output data. To obtain a unique order, a sinusoidal function inputs the position and the embedding dimensions. The self-attention module calculates the weights each input sequence assigns to other sequences. This way the self-attention reflects the dependencies between input and specific output and improves the modelling of long-range dependencies \cite{vaswani2017attention}. As such, the attention mechanism resembles a fully connected layer (FCL), whereby the weights reflect the pairwise relationship from previous inputs. To fully exploit this method, the transformer includes MHSA that splits the attention calculation among different heads of the embedding that can be calculated independently, facilitating its parallelization. Lastly, a feed-forward neural network (FFNN) is the subsequent layer to the MHSA. 

\subsection{Actor-Critic Reinforcement Learning}
RL teaches an agent to perform a task by accumulating experiences from interacting with its environment. Its actions are refined based on a reward function that signals the utility of these actions \cite{lillicrap2015continuous}. The RL's experimentation with the absence of prior knowledge mirrors the ignorance of the best combination of basic CNN layers. The only knowledge for the HPO process relates to the order of the stacked layer that starts with grid-like inputs and ends with an FCL. The combination of unknown orientation and trial-and-error experimentation matches the requirements of the HPO process. 

The RL methods can be split into value-based and policy-based methods. The value-based method estimates the quality of a state-action pair using a value function and optimizes this value iteratively using actions that maximize it. The high variability of value estimations and their under-performance in scenarios with continuous action spaces promotes policy-based methods \cite{lillicrap2015continuous}. The policy-based methods optimize the policy to maximize the expected cumulative rewards \cite{lillicrap2015continuous}. However, the latter methods suffer from sample inefficiency, which is an asset of value-based methods. To complement these methods' advantages, an actor-critic (AC) approach combines both policy-based (actor) and value-based (critic) methods \cite{lillicrap2015continuous}. Projected to the HPO problem, the actor outputs an action that maps to a layer and its HPs. The critic evaluates this action by outputting a value showing the action's quality \cite{lillicrap2015continuous}. 

\subsection{Transformers and Actor-critic RL}
The proposed framework, Transformer-based Reinforcement Learning HPO (\textit{TRL-HPO}) is the convergence of transformers and AC RL. The actor and the critic are implemented using the transformer's decoder architecture. The steps involved in the actions' generation, CNN model construction, and their evaluation are depicted in Figure \ref{fig:trhlpo_framework}. The combination of \textit{TRL-HPO} enables the RL agent to harness the strengths of transformers to benefit the HPO process. This new framework is the \textbf{first trial} to integrate transformers into the HPO process, opening a new frontier toward the exploitation of this novel architecture to address a lingering problem in the field of ML. This experiment evaluates the viability and suitability of this architecture to the HPO process. 

We analyze \textit{TRL-HPO} based on its inherent advantages and benefits versus the SOTA approaches. Two main advantages are reaped. The first is that \textit{TRL-HPO} builds models from scratch, rendering it a general-purpose method. The proof of concept is constrained to a single use case; however, the framework can be expanded to any DNN-related problem. The second concerns the definition of the reward function that is progressively updated with each stacked layer. While this definition incurs an extra computational footprint by training progressively deeper models, it facilitates understanding the contribution of each layer. This feature fits the requirements of environments that seek to balance resource constraints and accuracy objectives. 

The transformer architecture results in two main advantages to the HPO process. The first relates to handling long sequences, which benefits the transferability of \textit{TRL-HPO} from small to larger datasets. The second benefit relates to the MHSA, which facilitates the parallelization of \textit{TRL-HPO} and reduces its running time. Additionally, MHSA reveals the relationships between generated layers, which allows the inference of the combination of layers that produce better results, enhancing the transparency of the HPO process.

\section{Implementation Details}
The evaluation of \textit{TRL-HPO} is conducted on the MNIST dataset \cite{lecun1998mnist}, which is a large database of handwritten digits, from 0 to 9, containing 60,000 training and 10,000 testing images. Each image is a greyscale image of 28 $\times$ 28 dimensions. The availability and limited complexity of MNIST can serve as a good proof-of-concept for \textit{TRL-HPO}. The former factor enables a fair comparison with other SOTA approaches, such as \cite{baker2016designing, neary2018automatic}. The latter factor facilitates training CNN models in a limited time; an advantage when working with hardware constraints. The comparison with SOTA approaches is based on the convergence time to the best solution (hrs), the classification accuracy (\%), and the best CNN classification model (\%) obtained by each SOTA approach upon the completion of \textit{TRL-HPO} process (\textbf{AccTime}). 

Reproducibility is a major issue that plagues the HPO process \cite{elsken2019neural}. With this factor in mind, the choice of competing methods depended on two conditions: \textbf{(1)} The availability of source code that enables a fair comparison with \textit{TRL-HPO} and dispels any introduced biases by implementing a method from scratch. \textbf{(2)} The similarity in the experimental procedure for building CNNs from scratch using an RL-based method and the availability of results applied to the MNIST dataset. Therefore, we compare \textit{TRL-HPO} with Baker \textit{et al.} \cite{baker2016designing} and Neary \cite{neary2018automatic}, which abide by at least one of these conditions.
%
\begingroup
\setlength{\tabcolsep}{7pt}
\renewcommand{\arraystretch}{1.5}
\begin{table}[]
\centering
\begin{tabular}{|c|c|c|}
\hline
\textbf{Layer} & \textbf{HP} & \textbf{Values} \\ \hline
\multirow{3}{*}{\textbf{CNN}} & \textit{filters} & \{8, 16, ..., 128\} \\ \cline{2-3} 
 & \textit{kernel} & \{3, 5, 7\} \\ \cline{2-3} 
 & \textit{stride} & \{1, 2, 3\} \\ \hline
\multirow{3}{*}{\textbf{FCL}} & \textit{neurons} & \{16, 24, ..., 512\} \\ \cline{2-3} 
 & \textit{bias} & \{T, F\} \\ \cline{2-3} 
 & \textit{activations} & \begin{tabular}[c]{@{}c@{}}\{None, relu, leakyrelu, \\ tanh, sigmoid, elu, gelu\}\end{tabular} \\ \hline
\multirow{3}{*}{\textbf{MaxPool}} & \textit{kernel} & \{2, 3, ...,, 8\} \\ \cline{2-3} 
 & \textit{stride} & \{1, 2, 3\} \\ \cline{2-3} 
 & \textit{padding} & \{0, 1, 2, 3\} \\ \hline
\end{tabular}
\caption{Set of hyper-parameters}
\label{tab:HP}
\end{table}
\endgroup

The generated CNN models are constrained to 6 layers, formed as a combination of the three basic layers: CNN layer, FCL layer, and Maximum Pooling (MaxPool) layer. The set of HPs used in the experimental procedure is summarized in table \ref{tab:HP}. Stacking a combination of these layers adapts to the input structure. The state and action space design and the reward function are imperative to drive the RL agents' action generation, which requires defining the transformers' input sequence.

The reward function represents a layer's contribution towards improving the classification results, which means that the DNN model's first layer produces the highest rewards compared to any subsequent layers. Each layer is represented as a combination of the representation of the layer itself and its HP, and the performance of the obtained model on the MNIST's validation set. The generated layers' representation is obtained from the output layer of the RL's actor, summarized in four values. These values represent the action space mapped first to a layer and then to that layer's HPs. The model's performance is represented with 32 values, each calculating the model's accuracy results on each validation set's batch of size 16. The heterogeneity in the dimensions of the action space and the performance requires a mapping to a uniform representation. This goal is achieved using a static NN that takes these inputs and produces a uniform output of 64 values, referred to as Intermediate Model Representation (IMR). When a layer and its performance are obtained, the state space, reflecting the current state of the environment, should also change. Since the state space represents a sequence of layers, the index corresponding to the generated layer is updated with the IMR. With more layers, each index is updated with a new representation. This way the state space reflects the two important pieces of information for each layer, its HPs, and its performance. This process was touched upon in Figure \ref{fig:trhlpo_framework}.

The stopping criteria for the RL agent involve generating more than 6 layers, minimal improvement in the performance with the addition of layers (0.001), or accuracy below 60\%. These criteria are defined to avoid any unnecessary generation of models. A Deep Deterministic Policy Gradient (DDPG) \cite{lillicrap2015continuous} handles the continuous action spaces in this environment. The critic and the actor consist of target and online subnets to avoid radical changes in critic and actor updates \cite{lillicrap2015continuous}. In DDPG, the agent explores by adding random noise to the sampled actions, so that the agent can experiment with different combinations of layers and obtain their performance results. To remove data correlations for the RL agent inherent in the sequential structure of the data, experience replay (ER) \cite{lillicrap2015continuous} buffer is utilized to store data during the exploration stage.

The actor and the critic follow the transformer architecture, depicted in Figure \ref{fig:trhlpo_framework}. Two main differences highlight these components' distinctive roles: the transformer's output and the learning rate (lr). The actor's output is in the [0, 1] range mapped to a layer, while the critic's is in the [-1, 1] range representing the state's Q-value. Regarding the lr, it is recommended that the actors' (1e-5) be slower than the critics' (1e-4) \cite{lillicrap2015continuous}. The input embedding's dimension for each input space is equal to 64. The embeddings are inputs to multiple encoding layers, equivalent to 2 in our implementation, each constituting a Transformer block, represented by the MHSA and FFNN. Each of these blocks has the same number of input and output dimensions. The definition of MHSA and FFNN requires highlighting two values, \textbf{number of heads} equal to 4 and \textbf{expansion factor} equal to 4.

The training process is realized over several episodes. In each episode, 10 full models are generated via CPU parallelization. Once the ER buffer is full, five RL optimization rounds are implemented in each episode. The number of episodes and the size of the ER buffer are important parameters for the HPO process. Their optimization is imperative to obtain models with good performances in a short time. During the HPO process, the training set consists of 20,000 images while the validation set includes 10,000 images out of the original 60,000 images. The experimental procedure is conducted on SHARCNET's Graham cluster on a node with one V100 GPU, 64 Gbs of RAM, and 12 cores, such that each experiment does not exceed 7 days to avoid long queuing times. The implementation is available on the GitHub repository\footnote{https://github.com/Western-OC2-Lab/TRL-HPO}.  

\section{Results and Discussion}
This section reports and analyzes the results based on the defined performance metrics and layer-generation interpretability. \textit{TRL-HPO} results are obtained in the exploitation phase of the offline RL. 
\subsection{\textit{TRL-HPO} vs. SOTA}

\begingroup
\setlength{\tabcolsep}{6pt} 
\renewcommand{\arraystretch}{1.2}
\begin{table}[]
\centering
\begin{tabular}{|c|c|c|c|}
\hline
\textbf{Methods} & \textbf{\begin{tabular}[c]{@{}c@{}}Best Accuracy \\ (\%)\end{tabular}}& \textbf{\begin{tabular}[c]{@{}c@{}}Running Time\\ (hrs)\end{tabular}} & \textbf{\begin{tabular}[c]{@{}c@{}}AccTime\\ (\%)\end{tabular}} \\ \hline
\textit{TRL-HPO}           & 99.1                                                               & 99.3                                                                  & \textbf{94.6 \% $\pm$ 3.8\%}                                                \\ \hline
MetaQNN \cite{baker2016designing}          & \textbf{99.5}                                                               & 192-240                                                               & 88.5 \% $\pm$ 4.3\%                                                \\ \hline
Neary \cite{neary2018automatic}            & 95.8                                                               & \textbf{1.78}                                                                  & 95.8\%                                                          \\ \hline
\end{tabular}
\caption{\textit{TRL-HPO} versus SOTA in terms of accuracy and running time}
\label{tab:results_SOTA}
\end{table}
\endgroup

Table \ref{tab:results_SOTA} shows the results of \textit{TRL-HPO} versus SOTA methods that built CNN models from scratch. Based on evaluation results, \textbf{MetaQNN} produces the best model when classifying MNIST data reported in the \textbf{exploration phase}. This result is due to the controller's freedom to generate models up to 12 layers and the integration of priors by adding dropout layers \cite{baker2016designing}. The accumulation of more layers improves classification results; however, at the expense of wider search space that increases the running time. This is evident with the computational footprint of \textbf{MetaQNN} that exceeds \textit{TRL-HPO}, despite running on 8$\times$ the amount of GPU resources. On the other hand, the approach in \cite{neary2018automatic} is the first to converge based on the reported results. However, this method experimented with four configurations, highlighting the limited exploration of this strategy and the pre-mature convergence of the master agent. This limitation is reflected by its poor best models compared to the other two approaches, \textbf{excluding} it from further analysis. Since the larger search space of \textbf{MetaQNN} would not provide a fair comparison with \textit{TRL-HPO}, it was imperative to compare these two approaches using \textbf{AccTime}. Within \textit{TRL-HPO} convergence time, its generated models exhibit the best performance compared to the SOTA models. Despite \textit{TRL-HPO}'s need to generate more models given the requirement of progressive reward, it outperforms other models in the \textbf{AccTime} metric. Compared to other SOTA methods, \textit{TRL-HPO} can progressively generate CNN models with satisfactory performance in a shorter period. These two conditions are important for resource-constrained IoT environments. In terms of \textbf{AccTime}, on average, \textit{TRL-HPO} outperforms \textbf{MetaQNN} by 6.8 \% in accuracy.   The improvement in the accuracy result of \textit{TRL-HPO} compared to \textbf{MetaQNN} is statistically significant (p-value = 0.02) by applying a t-test on the distribution of \textit{TRL-HPO} versus the 1-sd values of \textbf{MetaQNN}.

In a resource-constrained IoT environment, analyzing the energy 
consumption and processing power of \textit{TRL-HPO} compared to SOTA methods is imperative. We analyze these methods using the method's number of parameters and the floating point operations (FLOPs), which cover IoT-related concerns. The \textbf{MetaQNN} requires storing all models and state transitions to find the best-performing models, which with many combinations of layers and HPs, is prohibitive for the IoT environment. In Neary \textit{et al.} approach \cite{neary2018automatic}, an RL agent is assigned to every HPs and a master agent finds the best combination of individual HPs. For Neary \textit{et al.}, 399K FLOPs are required for inference whereas \textit{TRL-HPO} requires 991k FLOPs. On the other hand, both models have the same number of model parameters (78.8k parameters).

\subsection{Layer Analysis}
Investigating the effect of the addition of layers is central to understanding the contribution of each layer towards improving accuracy results and the combination of layers that yields the best performances. This analysis unveils the complex relationships between layers facilitating the transparency of the HPO process. Both goals were considered in the design of \textit{TRL-HPO}, representing a key differentiating factor versus SOTA methods. To gain these insights, two questions need to be answered. \textbf{(1)} What are the layer combinations degrading the performance of CNN models? and \textbf{(2)} How are layer affinities reflected in the attention mechanism? 

\begin{figure}
    \centering
    \includegraphics[scale=0.4]{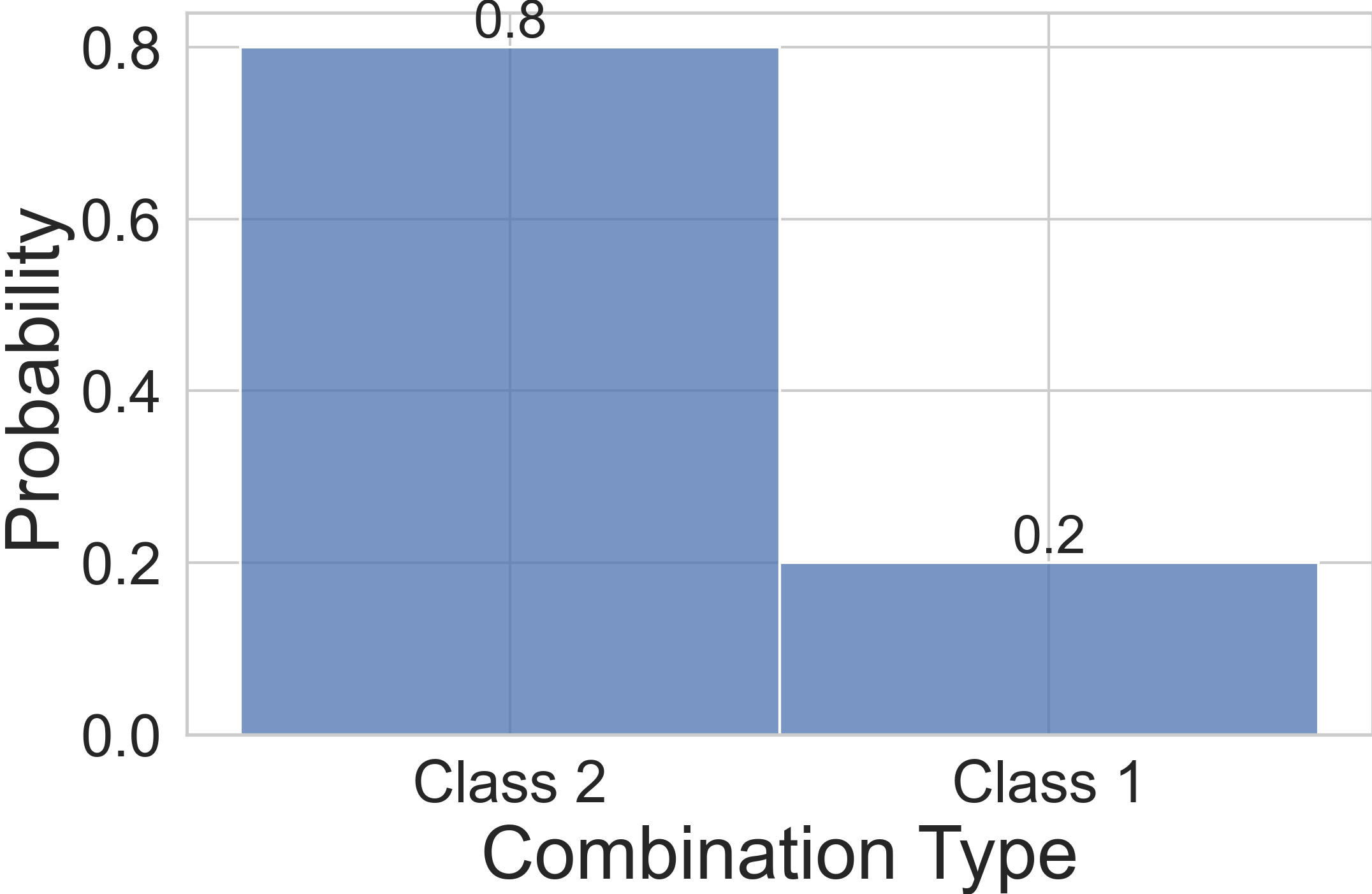}
    \caption{Layer affinity for models with negative rewards}
    \label{fig:neg_rewards}
\end{figure}

Figure \ref{fig:neg_rewards} summarizes the distribution of layer combinations with negative rewards. Two combinations stand out, \textbf{Class 1} represented by Conv2D and Conv2D combinations and \textbf{Class 2} represented by two consecutive FCLs. A layer combination refers to the layer that produced a negative reward and its previous layer. Two main insights can be gathered from Figure \ref{fig:neg_rewards}. The first is that the negative reward is overwhelmingly attributed to the accumulation of FCLs. This means that the stacking of multiple FCL layers is superfluous on the MNIST dataset, a conclusion that aligns with the best models obtained in \cite{baker2016designing}. The second insight highlights that additional Conv2D layers can degrade model performance, especially with MNIST data of limited visual complexity. As such, these observations are beneficial to garner knowledge about stacking of layers, suitable for environments with resource constraints. 

\begin{figure}
    \centering
    \includegraphics[scale=0.4]{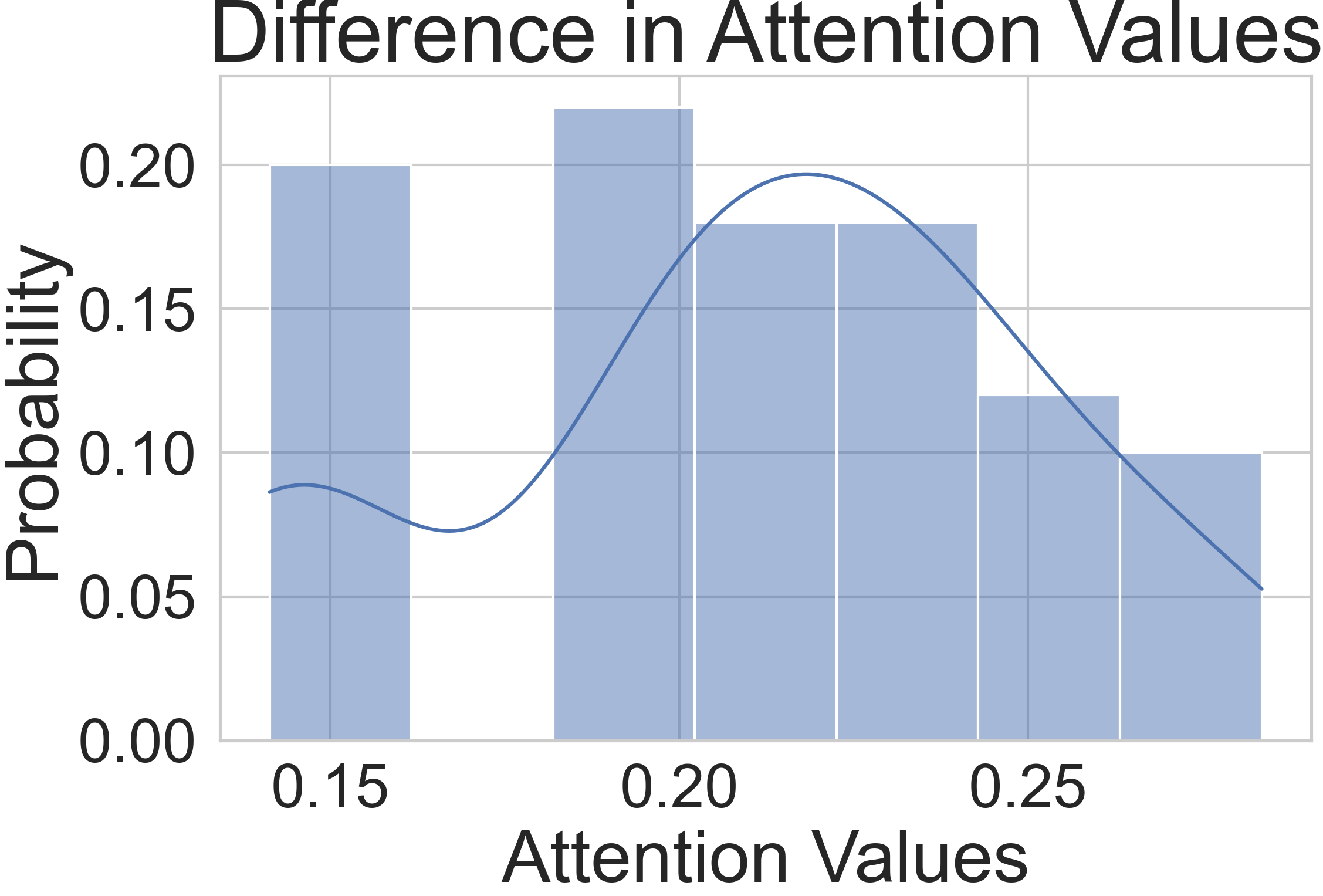}
    \caption{Differences in attention between layers of best reward and layers with negative reward}
    \label{fig:diff_attn}
\end{figure}

Figure \ref{fig:diff_attn} depicts the distribution of the differences in attention values between layers with the best performances and layers with negative rewards. The distribution demonstrates the close fidelity of the attention mechanism and the utility of a specific layer in its classification performance. As such, the attention mechanism captures the close affinity between layers assigning higher weights to promising layers while shunning layers with poor performances. The difference in attention values averages 0.22, a significant value considering that some attention is assigned to the layers yet to be generated. In these cases, the attention values of the layer degrading model performances are higher than these to-be-generated layers. Therefore, we conclude that the attention mechanism reflects the relationship between layers within the context of generating best-performing models. 

\begingroup
\setlength{\tabcolsep}{7pt}
\renewcommand{\arraystretch}{1.7}
\begin{table*}
    \centering
    \begin{tabular}{| m{3cm} | m{5cm} | m{5cm} |}
        \hline
        \multicolumn{1}{|c|}{\textbf{Challenge}} & \multicolumn{1}{|c|}{\textbf{Summary}} & \multicolumn{1}{|c|}{\textbf{Solutions}}\\ 
        \hline
        Computational Time  & 
        \begin{itemize}[leftmargin=*]
            \item[--] Training of individual models
            \item[--] Optimization rounds of actor-critic approach
        \end{itemize} 
        & 
        Storing the trained models in a hash table to avoid re-training
        \\
        \hline 
        Exploration Strategy & 
        DDPG's lack of direction & 
        Application of RL methods with inherent randomness such as soft actor-critic approach
        \\
        \hline 
         Reward Function & 
         \begin{itemize}[leftmargin=*] 
            \item[--] Training many models to infer their contribution to classification results
            \item[--] Prioritization of accuracy
        \end{itemize}
         & 
        \begin{itemize}[leftmargin=*] 
            \item[--] Generating a block of similar layers
            \item[--] Inclusion of inference time, energy consumption, and number of operations per second
        \end{itemize}
        \\
        \hline
         Reward Function & 
         \begin{itemize}[leftmargin=*] 
            \item[--] Training many models to infer their contribution to classification results
            \item[--] Prioritization of accuracy
        \end{itemize}
         & 
        \begin{itemize}[leftmargin=*] 
            \item[--] Generating a block of similar layers
            \item[--] Inclusion of inference time, energy consumption, and number of operations per second
        \end{itemize}
        \\
        \hline
        Generalizability, Scalability, and Transferability & 
         \begin{itemize}[leftmargin=*] 
            \item[--] Scalability and Transferability concerns with larger datasets and complex architectures.
            \item[--] Generalizability to different IoT Environments
        \end{itemize}
         & 
        \begin{itemize}[leftmargin=*] 
            \item[--] Expansion of the input sequence and actor's block generation
            \item[--] Changing the mappings of the actor's output to adapt to different DNN models
        \end{itemize}
        \\
        \hline
        Integration into Automated ML & 
         N/A
         & 
        Replacing the HPO process for IoT deployment frameworks
        \\
        \hline
    \end{tabular}
    \caption{Summary of Challenges}
    \label{tab:challenges_summary}
\end{table*}
\endgroup
 
\section{Open Challenges and Future Recommendations}
The implementation of the \textit{TRL-HPO} process opens new avenues for exploring the improvement of this process, which spans many research and practical questions about RL-based HPO implementations, summarized in Table \ref{tab:challenges_summary}. 

\subsection{Computational Time}
Despite improvements in the efficiency of the \textit{TRL-HPO} process, it has yet to converge in a short time. The bottleneck in the training process originates from two main sources: \textbf{(1)} the training of individual models, and \textbf{(2)} optimization rounds of the RL approach. The first source is a necessary evil because of the exploration requirements and the need to evaluate the actor's actions. Determining the training data size, the number of epochs, and learning rates are required to achieve close to optimal accuracy. During the exploration and action assessment phases, repetitive models can be generated. Storing these models can avoid retraining them, which can overwhelm the available GPU or RAM resources, especially with a large ER buffer. An alternative method is storing already trained models in a database or a hash table, whereby the model is represented via its HP or its hash value. A salient issue relates to querying algorithms that should search for HP representations available in the database. The querying time will increase with the expansion of the database, requiring a more intricate search procedure. These methods should be investigated to budget the computational time towards more fruitful procedures that benefit RL-based solutions. 

\subsection{Exploration}
RL implementations depend on the trial-and-error procedure in the exploration phase. However, a shortcoming of this phase, either through DDPG's random noise or $\epsilon$-greedy algorithms \cite{baker2016designing}, relates to its lack of direction. This means that the agent continues exploring unpromising areas of the search space with relatively poor performances. Therefore, it is imperative to include priors or rectify the exploration stage to reduce unnecessary exploration. Aspects of the BO process should be integrated into the RL-based strategies to streamline the exploration process. The randomness can be incorporated using a stochastic policy, which is part of the soft actor-critic policy gradient approach \cite{haarnoja2018soft}. In the exploration stage, the stochastic policy outputs the normal distribution's mean and standard deviation. The standard deviation is progressively reduced to achieve a deterministic policy based on the reward function. This suggestion opens new frontiers toward examining methods that can associate the uncertainty of the stochastic policy with steering the RL's agent exploration.   

\subsection{Reward Function}
The progressive reward function of \textit{TRL-HPO} is foundational to gaining insights into the contribution of each layer to performance enhancement. However, this layer accumulation process requires training many models to infer their performance, restricting the model generation process to a few layers. These constraints can be sidestepped by generating similar layers at once, which is referred to as a block generation procedure. As such, computational time is gained at the expense of transparency and layer diversity. In resource-constrained environments, factors such as energy consumption, inference time, and number of computational operations per second can gain precedence over accuracy. These factors can be added to the reward function and assigned a weight based on application requirements.

\subsection{Generalizability, Scalability, and Transferability}

With each contribution toward the HPO procedure, questions of generalizability, scalability and transferability loom to undermine their usefulness; a concern that applies to RL-based solutions. In \textit{TRL-HPO}, the actor and the critic generate and evaluate models using FCL, CNN, and MaxPool layers, necessary to build DNN models. When applied to complex datasets and architectures, skip and residual connections can be integrated into the \textit{TRL-HPO} architecture. Towards that end, two main approaches can be followed: (1) expand the input space to include more layers, which is reflected in the sequence length variable, or (2) change the definition of the actor’s output from the layer and its HPs to include the number of similar layers and their HPs. Adopting any of these approaches should consider the trade-off between the availability of computational resources and the layer generation process’s transparency. The same adaptation procedure can be followed when confronting a transferability concern.

The IoT environment is integrated into various fields, including smart cities, healthcare, buildings, and electric grids. This environment suffers from computational and communication resource scarcity, which can be detrimental to ML applications, such as anomaly detection, object detection, and forecasting. In connection to the IoT environment limitations, the \textit{TRL-HPO} framework offers solutions manifested in three aspects: (1) its generalizability to different types of DNN models, including CNN, Long short-term memory (LSTM) and FFNN by changing the mappings of the actor's output, (2) its low computational, storage, and processing footprint, compared to SOTA approaches, and (3) its transparency that enables the reliance on part and not all of the CNN models and exchanging model parameters when the need arises.

\subsection{Integration into AutoML}
The HPO process is an important procedure in the AutoML pipeline and its enhancement is central to the wide-scale deployment and adoption of these pipelines. The \textit{TRL-HPO} process is envisioned to replace the generic HPO processes that are part of the ubiquitous cloud computing ML deployment modules. These modules enhance the efficiency and accessibility of ML deployment, by providing user-friendly deployment strategies. \textit{TRL-HPO} that outperformed SOTA approaches in its convergence time while maintaining good accuracy results benefits experts who have budgetary constraints and desire to prove the viability of their products/models for customers/investors alike.

\section{Conclusion}
The HPO process is a fundamental step in the ML pipeline that enhances model performance. However, its computational footprint is prohibitive for widespread adoption and current methods overlook the transparency in the layer generation process. These factors are imperative to realize IoT applications' function, AVs in particular, requiring model partitioning to fulfill the resource constraints of edge environments. To address these limitations, this paper proposes \textit{TRL-HPO} framework that combines transformers with an RL actor-critic approach. The attention mechanism, parallelization, and the progressive generation of layers are all novel properties of this framework within the transparency and time requirements. These advantageous factors were empirically established compared to SOTA approaches using the MNIST dataset. Moreover, a list of research questions scrutinizing RL-based solutions, including \textit{TRL-HPO}, is presented with corresponding recommendations. Future work will target the presented questions to design a general-purpose HPO process. 

\section{Acknowledgement}
This work was made possible by the facilities of the Shared Hierarchical 
Academic Research Computing Network (SHARCNET:www.sharcnet.ca) and Compute/Calcul Canada.

\bibliographystyle{IEEEtran}
\bibliography{refs}

\end{document}